\documentclass[a4paper]{article}
\usepackage{amsmath,graphicx, subfig, subcaption, float}
\usepackage[normalem]{ulem}
\usepackage{hhline}
\usepackage{authblk}

\usepackage{multirow}
\usepackage{hyperref}
\usepackage[table,xcdraw]{xcolor}

\usepackage{enumitem}
\setlist{nosep, leftmargin=14pt}

\title{Ensemble of Small Classifiers For Imbalanced White Blood Cell Classification}
\date{}
\author[1,$\star$]{Siddharth Srivastava}
\author[1]{Adam Smith}
\author[2,3]{Scott Brooks}
\author[1]{Jack Bacon}
\author[1]{Till Bretschneider}
\affil[1]{Department of Computer Science, University of Warwick}
\affil[2]{Warwick Medical School, University of Warwick}
\affil[3]{Intelligent Imaging Innovations Ltd}
\affil[$\star$]{Corresponding author: siddharth.srivastava@warwick.ac.uk}

\begin{document}
\maketitle
\begin{abstract}
Automating white blood cell classification for diagnosis of leukaemia is a promising alternative to time-consuming and resource-intensive examination of cells by expert pathologists. However, designing robust algorithms for classification of rare cell types remains challenging due to variations in staining, scanning and inter-patient heterogeneity. We propose a lightweight ensemble approach for classification of cells during Haematopoiesis, with a focus on the biology of Granulopoiesis, Monocytopoiesis and Lymphopoiesis. Through dataset expansion to alleviate some class imbalance, we demonstrate that a simple ensemble of lightweight pretrained SwinV2-Tiny, DinoBloom-Small and ConvNeXT-V2-Tiny models achieves excellent performance on this challenging dataset. We train $3$ instantiations of each architecture in a stratified $3$-fold cross-validation framework; for an input image, we forward-pass through all $9$ models and aggregate through logit averaging. We further reason on the weaknesses of our model in confusing similar-looking myelocytes in granulopoiesis and lymphocytes in lymphopoiesis. Code: \url{https://gitlab.com/siddharthsrivastava/wbc-bench-2026}

\end{abstract}
\section{Introduction}
Haematological image analysis is clinically important for the early diagnosis of Leukaemia, a group of cancers originating in the bone marrow, which lead to an overproduction of abnormal white blood cells (WBC). Traditional diagnosis often relies on microscopic examination performed by a human expert, which is resource intensive and time consuming. Automatic WBC classification promises a low resource alternative for the detection of rare blood cells. Some methodologies are emerging, but still require further validation before they can be put to clinical use, specifically in terms of their robustness \cite{sadafi2023continuallearningapproachcrossdomain}. The ISBI WBCBench 2026: Robust White Blood Cell Classification challenge \cite{wbcbench2026} was created to benchmark machine learning models for classifying white blood cells into 13 classes. These classes represent the development of WBCs through Haematopoiesis along three major branches that produce the broader classes of granulocytes, monocytes and lymphocytes. \cite{jagannathan2013hematopoiesis}. Despite advances in deep learning models for cell classification \cite{shifat2020cell}, this problem remains challenging due to inherent limitations in training data, such as noise, stain variation and class imbalance. To address this we propose an ensemble of lightweight classifiers to achieve near state-of-the-art performance with limited compute.

\section{Dataset}
\subsection{Dataset Expansion}
\label{section:dataset_expansion}

\begin{figure}[t]
    \centering
    \includegraphics[width=0.95\linewidth]{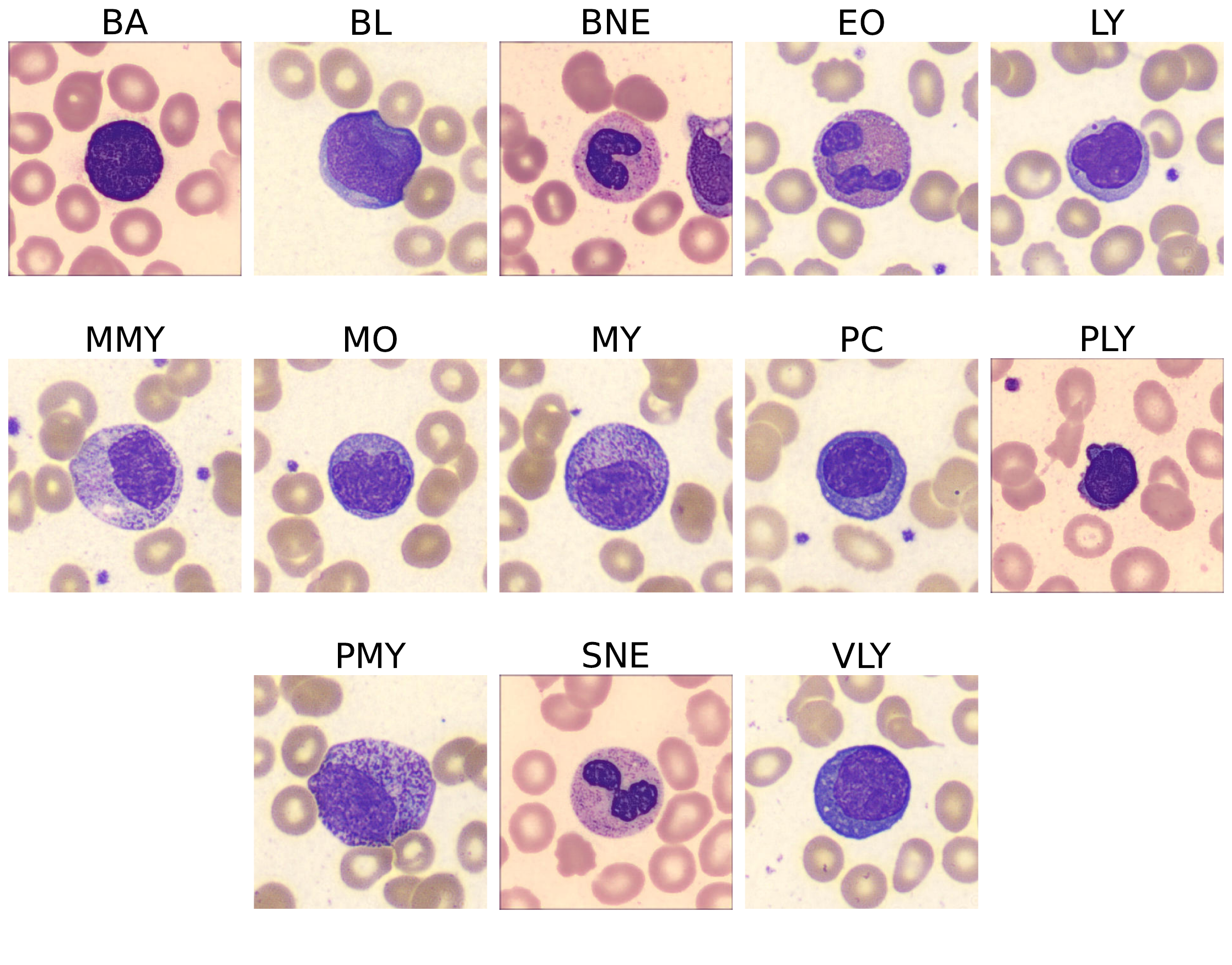}
    \caption{Example phase 1 (unperturbed) images for each class from the WBCBench competition dataset.}
    \label{fig:example_images}
\end{figure}

\begin{figure}[t]
    \centering
    \includegraphics[width=0.95\linewidth]{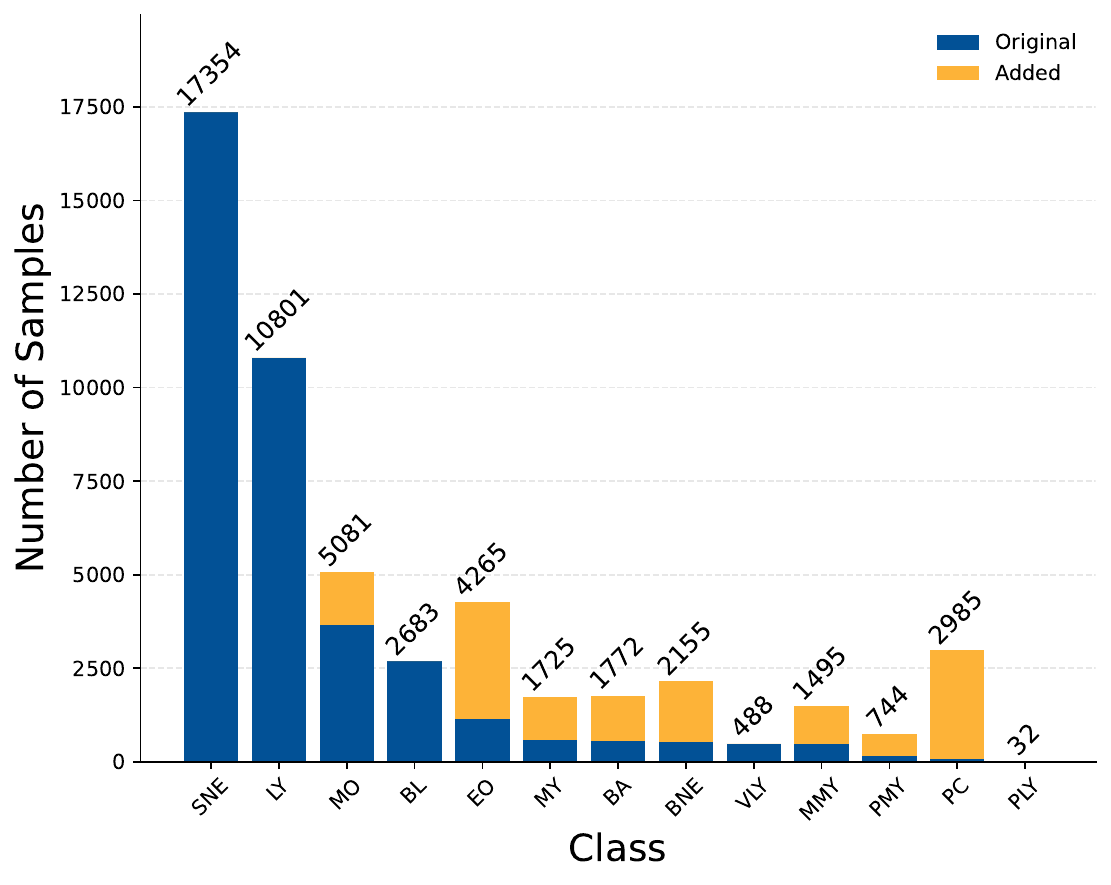}
    \caption{Number of samples per class in our expanded dataset.}
    \label{fig:expanded_dataset}
\end{figure}

The challenge dataset (WBCBench \cite{wbcbench2026}) consists of $55,012$  H\&E stained peripheral blood smear images of size $368 \times 368$ across 13 classes, capturing all cell types found during Haematopoiesis: segmented neutrophils (SNE), lymphocytes (LY), monocytes (MO), blast cells (BL), eosinophils (EO), myelocytes (MY), band-form neutrophils (BNE), variant (atypical) lymphocytes (VLY), metamyelocytes (MMY), promyelocytes (PMY), plasma cells (PC), and prolymphocytes (PLY); Figure \ref{fig:example_images} showcases example images from each class. $16,477$ images are unlabelled and reserved for testing, leaving $38,535$ annotated images for model training. The dataset is highly imbalanced, with SNE and LY classes together comprising $28,155$ of the images, and the two least-represented classes, PC and PLY, only $90$ and $14$, respectively. Therefore, we expanded the dataset using external public datasets:
\begin{itemize}
    \item\textit{Acevedo-20} \cite{ACEVEDO2020105474}: Over $17,000$ images, of which we use $10,312$ images from the following classes: MO (1420), EO (3117), BNE (1633), PMY (592), BA (1218), MY (1137) and MMY (1015);
    \item\textit{Blood\_dataset} by Taeyeon Kim \cite{huggingfaceEhottlblood_datasetDatasets}, originally developed as the Blood 8 classes dataset \cite{blood_8_classes_dataset}: Over $46,000$ images from 8 classes, of which we only use $2895$ plasma-cell images. 
    \item Prolymphocyte images from CellWiki \cite{cellwikiProlymphocyteCellWiki}: $18$ additional prolymphocyte examples bring the total number of PLY instances to $32$. 
\end{itemize}

In total, we include additional $13,045$ training examples, alleviating some of the most severe class imbalances in the original dataset (per-class frequencies shown in Figure \ref{fig:expanded_dataset}). We showcase that adding external public data results in a more robust model in Table \ref{table:ablation}.

To emulate real-world domain shift in scanner and stain variability,  WBCBench introduces augmentations in the form of Gaussian noise, motion blur and colour perturbations. While motion blur and colour perturbations can be addressed through train time augmentations, we tackle noise degradation adaptively with non-local means denoising \cite{Buades2005} estimating  Gaussian noise using a robust wavelet-based estimator for the standard deviation, $\hat{\sigma}$ \cite{Donoho1994}. The patch distance for denoising is set to $\sqrt{\hat{\sigma}}$. We empirically find this setup retains the fine-grained morphological detail of clean images while sufficiently denoising the noisy images. The extreme class imbalance in the training dataset necessitates the use of a \emph{weighted} sampler when sampling batches for training, such that rare classes are sampled at the same rate as the most common classes. Thus, we weight each class based on the \emph{effective number} $E_n$ \cite{cui2019classbalancedloss}:

\[ E_n = (1 - \beta^n) / (1 - \beta),\]

$n$ being the class size, and $\beta = 0.9999$ a hyperparameter.

\section{Method}

\begin{figure*}[ht]
    \centering
    \begin{minipage}{0.49\linewidth}
    \centering
    \includegraphics[width=\linewidth]{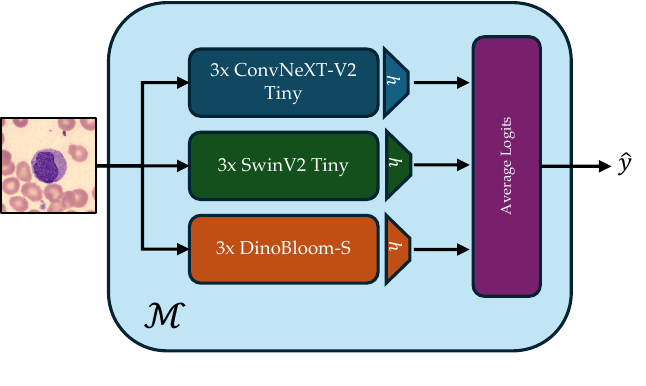}
    \end{minipage}
    \hfill\vline\hfill
    \begin{minipage}{0.45\linewidth}
    \centering
    \includegraphics[width=\linewidth]{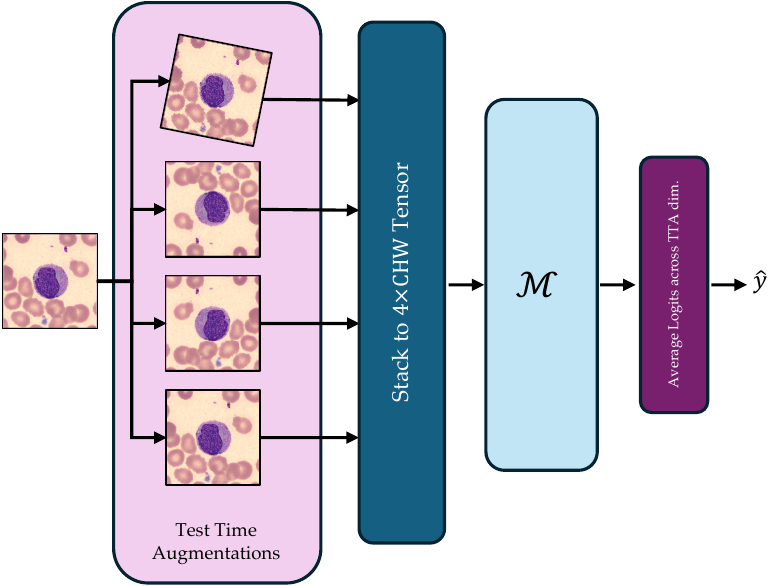}
    \end{minipage}
    \caption{Model architecture. \textbf{Left:} our model consists of a $3$-ensemble of $3$ different architectures: a ConvNeXT-V2 Tiny \cite{10205236}, a SwinV2-Tiny \cite{Liu_2022_CVPR}, and a DinoBloom-Small \cite{Koc_DinoBloom_MICCAI2024} model. In each, we replace the pretrained classification head with a simple MLP $h$. Each instantiation of an architecture is fined-tuned independently on $2$ out of $3$ folds of data and validated on the last fold to mitigate overfitting, giving 3 models for each architecture and a total of $9$ models. The final model $\mathcal{M}$ passes an input image through all $9$ architectures and averages the logits to produce the $13$-class logit vector. \textbf{Right}: At inference, we generate multiple augmented views of each image via random flips and rotations. These views are independently processed by $\mathcal{M}$ and their logits are averaged to obtain the final class prediction. }
    \label{fig:model_architecture}
\end{figure*}

\subsection{Model Architecture}
We outline our model architecture in Figure \ref{fig:model_architecture}, which consists of instantiations of $3$ pretrained classifiers highlighted below:
\begin{itemize}
    \item Swin Transformer V2 \cite{Liu_2022_CVPR}: a hierarchical vision transformer whose representation is composed of non-overlapping shifted windows. Swin-V2 improves upon the original Swin transformer architecture \cite{liu2021Swin} by introducing techniques to stabilise training, better handling of differing image resolutions and a new pretraining method to reduce the amount of training data. We use \texttt{swinv2-tiny}, which consists of 27.5M parameters, and replace the classification head with a small MLP, giving a total of $28.1$M parameters. 
    \item ConvNeXT-v2 \cite{10205236}: a pure convolutional neural network that improves upon ResNets by using LayerNorm layers in place of BatchNorm layers, GELU instead of ReLU, and fewer activation functions \cite{liu2022convnet}. ConvNeXT-V2 improves upon the original ConvNeXT by using a fully convolutional masked autoencoder framework and a global response normalisation layer \cite{10205236}. We use \texttt{convnextv2-tiny} and similarly replace the classification head with a MLP, giving a total of $28.4$M parameters. 
    \item DinoBloom \cite{Koc_DinoBloom_MICCAI2024}: a foundation model for single-cell haematology images built upon DINOv2 \cite{oquab2023dinov2}.  We use \texttt{DinoBloom-Small} with our own classification head, totalling $23$M parameters.
\end{itemize}
The selected backbone models provide complementary inductive biases: SwinV2 is a transformer-based model using the hierarchical attention mechanism and shifted windows, which enables interactions and reasoning between long-range patches in an image. In contrast, ConvNeXT-V2 retains the built-in locality and translation equivariance biases of convolutional neural networks \cite{liu2022convnet}, which is effective for fine-grained texture and morphological patterns. Lastly, DinoBloom is a domain-specific foundation model for haematological analyses, unlike ConvNeXT and Swin which are trained on natural images \cite{Liu_2022_CVPR, 10205236}, providing representations tailored to cellular morphology and H\&E stains. The incorporation of a wide variety of networks with different biases and training data provides a rich representation of our images for classification. All MLP classification heads consist of 4 blocks of Linear-ReLU-Dropout ($p=0.1$)

\subsection{Model Training}
We partition the expanded training dataset (section \ref{section:dataset_expansion}) into $3$ stratified folds and train $3$ independent instances of each architecture, each using two folds for training and the remaining fold for validation to limit overfitting. For each model, we use the $\alpha$-balanced focal loss \cite{Lin2017focalloss}, with $\alpha$ and $\gamma$ set to $0.25$ and $2$, respectively. Focal loss reduces the relative loss on well-classified examples with high $p_t$ and shifts the focus onto hard, misclassified examples with low $p_t$. We train using batch size $32$ using the AdamW optimiser \cite{loshchilov2019decoupledweightdecayregularization}, with $\beta = (0.9, 0.999)$ and a cosine-decaying learning rate starting at $5\times10^{-4}$ and decaying to $5\times10^{-6}$, no LR warmup, and a patience of $10$ epochs on macro-F1 for the validation set to stop training. We use an exponential moving average (EMA) of the model weights with $\beta$ set to $0.999$. Lastly, we employ image augmentations to emulate the corruption in the training data and improve robustness: random flips, rotations, Gaussian noise and blurring, motion blurring, and random brightness and contrast. Further, we use cutmix \cite{yun2019cutmixregularizationstrategytrain} and mixup \cite{zhang2018mixupempiricalriskminimization}, each with $0.15$ probability per batch. With full parameter fine-tuning, each model trains at 5 minutes per epoch on 3 NVIDIA A10 cards in parallel.

\subsection{Model Inference}
For inference we perform test-time augmentation, shown in Figure \ref{fig:model_architecture} right; we pass a test image through a series of augmentations (random flips and random rotation), pass those images through the model and average the output logits to obtain a `smooth' logit representation of the original input image. A full pass through the test dataset of $16,447$ images with our $9$ models takes $30$ minutes on a single NVIDIA A40.

\begin{figure*}[ht!]
    \centering
    \begin{minipage}{0.49\linewidth}
    \centering
    \includegraphics[width=\linewidth]{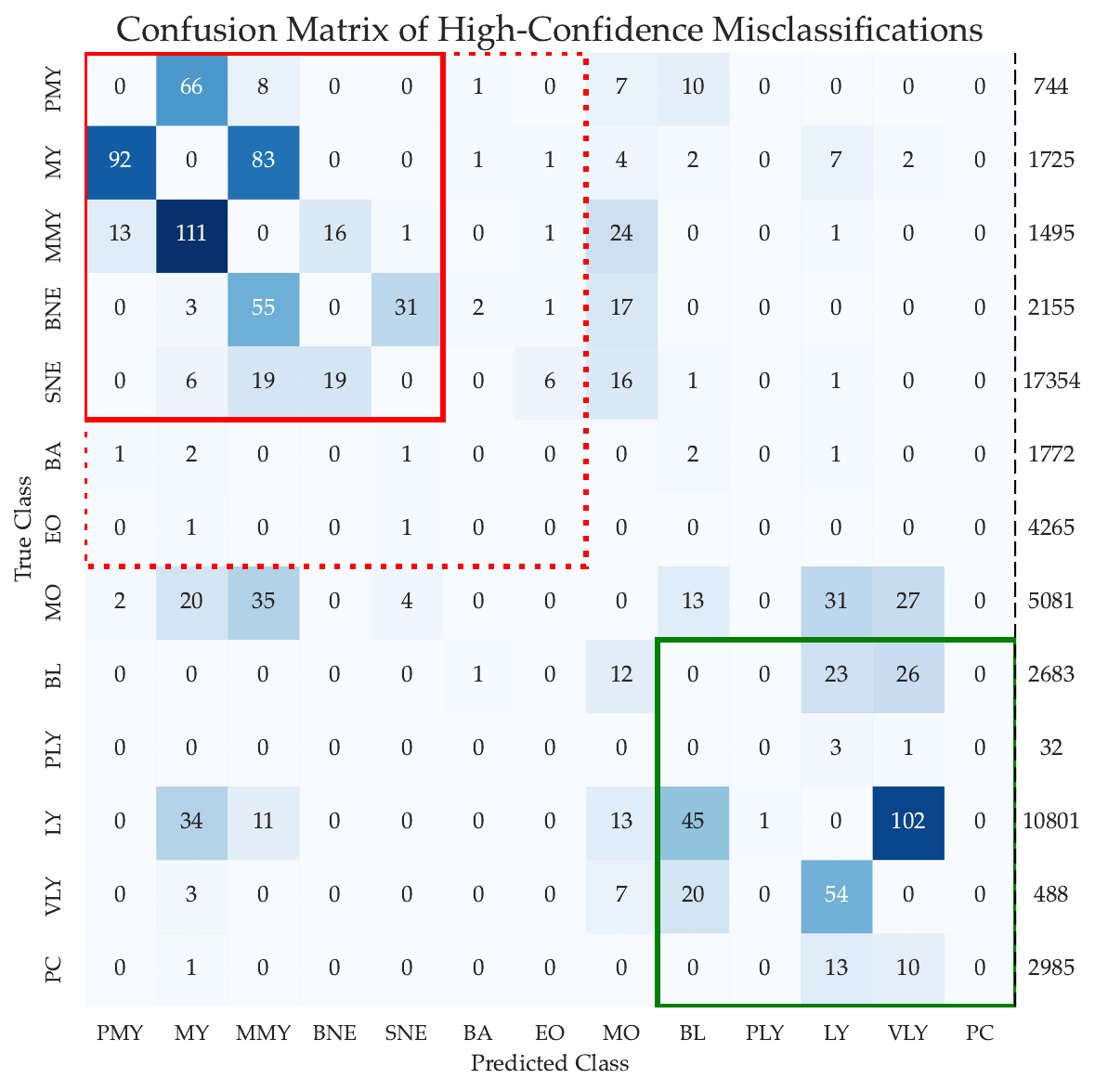}
    \end{minipage}
    \begin{minipage}{0.49\linewidth}
    \centering
    \includegraphics[width=0.97\linewidth]{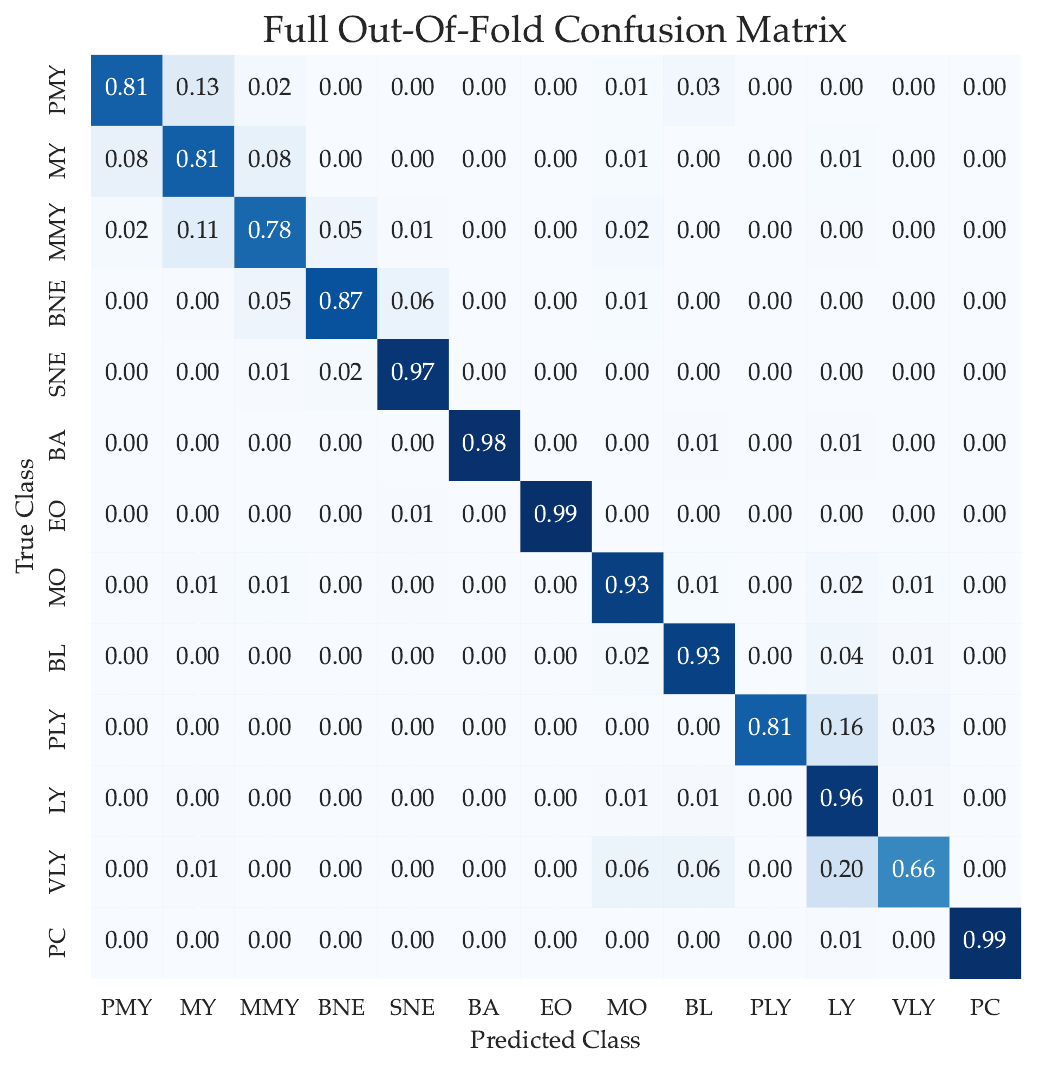}
    \end{minipage}
    \caption{Confusion matrices of our ensemble trained on the expanded dataset. \textbf{Left}: confusion matrix of $1149$ high-confidence misclassified examples, $841$ of which come from WBCBench and the rest from Acevedo20; found using confident learning \cite{northcutt2021confidentlearning}. The dotted red box groups the classes in granulopoiesis and the green box captures lymphopoiesis; MO occurs in monocytopoiesis. \textbf{Right} confusion matrix generated using out-of-fold examples for each group of models. }
    \label{fig:confusion_matrices}
\end{figure*}

\useunder{\uline}{\ul}{}
\begin{table*}[h!]
\centering
\renewcommand{\arraystretch}{1.0}
\resizebox{\textwidth}{!}{%
\begin{tabular}{cc|ccccc||ccccc}
\multicolumn{2}{c|}{\cellcolor[HTML]{EFEFEF}} & \multicolumn{5}{c||}{Training Dataset: WBCBench Train + Eval} & \multicolumn{5}{c}{Training Dataset: Expanded Dataset} \\ \hhline{~~----------}
\multicolumn{2}{c|}{\multirow{-2}{*}{\cellcolor[HTML]{EFEFEF}}} & \multicolumn{4}{c}{\begin{tabular}[c]{@{}c@{}}Eval. Set: WBCBench Train + Eval\\ Out-of-Fold\end{tabular}} & \cellcolor[HTML]{FFECF0}\begin{tabular}[c]{@{}c@{}}WBCBench \\ Test\end{tabular} & \multicolumn{4}{c}{\begin{tabular}[c]{@{}c@{}}Eval. Set: WBCBench Train + Eval \\ Out-of-Fold\end{tabular}} & \cellcolor[HTML]{FFECF0}\begin{tabular}[c]{@{}c@{}}WBCBench \\ Test\end{tabular} \\ \hhline{~~----------}
\multicolumn{1}{c|}{} & Model & \begin{tabular}[c]{@{}c@{}}Macro\\ F1\end{tabular} & \begin{tabular}[c]{@{}c@{}}Balanced\\ Accuracy\end{tabular} & \begin{tabular}[c]{@{}c@{}}Macro\\ Precision\end{tabular} & \begin{tabular}[c]{@{}c@{}}Macro\\ Sensitivity\end{tabular} & \cellcolor[HTML]{FFECF0}\begin{tabular}[c]{@{}c@{}}Macro \\ F1\end{tabular} & \begin{tabular}[c]{@{}c@{}}Macro\\ F1\end{tabular} & \begin{tabular}[c]{@{}c@{}}Balanced\\ Accuracy\end{tabular} & \begin{tabular}[c]{@{}c@{}}Macro\\ Precision\end{tabular} & \begin{tabular}[c]{@{}c@{}}Macro\\ Sensitivity\end{tabular} & \cellcolor[HTML]{FFECF0}\begin{tabular}[c]{@{}c@{}}Macro\\ F1\end{tabular} \\ \hline
\multicolumn{1}{c|}{} & $3 \times$SwinV2 & 0.7351 & 0.7759 & 0.7126 & 0.9932 & \cellcolor[HTML]{FFECF0} & 0.7410 & 0.7523 & 0.7412 & 0.9936 & \cellcolor[HTML]{FFECF0} \\
\multicolumn{1}{c|}{} & $3 \times$ConvNeXt & 0.7523 & 0.7669 & 0.7488 & 0.9938 & \cellcolor[HTML]{FFECF0} & 0.7662 & 0.7660 & 0.7741 & 0.9942 & \cellcolor[HTML]{FFECF0} \\
\multicolumn{1}{c|}{} & $3 \times$DinoBloom & 0.7276 & {\ul 0.8095} & 0.6863 & 0.9925 & \multirow{-3}{*}{\cellcolor[HTML]{FFECF0}-} & 0.7417 & \textbf{0.7739} & 0.7296 & 0.9936 & \multirow{-3}{*}{\cellcolor[HTML]{FFECF0}-} \\
\multicolumn{1}{c|}{\multirow{-4}{*}{\begin{tabular}[c]{@{}c@{}}No\\  TTA\end{tabular}}} & Ensemble & {\ul 0.7666} & 0.789 & {\ul 0.7586} & {\ul 0.9941} & \cellcolor[HTML]{FFECF0}{\ul 0.6638} & {\ul 0.7791} & {\ul 0.7736} & {\ul 0.7977} & \textbf{0.9945} & \cellcolor[HTML]{FFECF0}{\ul 0.6737} \\ \hline
\multicolumn{1}{c|}{} & $3 \times$SwinV2 & 0.7477 & 0.7806 & 0.7306 & 0.9935 & \cellcolor[HTML]{FFECF0} & 0.7499 & 0.7503 & 0.7666 & 0.9939 & \cellcolor[HTML]{FFECF0} \\
\multicolumn{1}{c|}{} & $3 \times$ConvNeXt & 0.7557 & 0.7688 & 0.7535 & 0.9939 & \cellcolor[HTML]{FFECF0} & 0.7724 & 0.7696 & 0.7815 & {\ul 0.9944} & \cellcolor[HTML]{FFECF0} \\
\multicolumn{1}{c|}{} & $3 \times$DinoBloom & 0.7323 & \textbf{0.8166} & 0.6901 & 0.9928 & \multirow{-3}{*}{\cellcolor[HTML]{FFECF0}-} & 0.7419 & 0.7718 & 0.7321 & 0.9939 & \multirow{-3}{*}{\cellcolor[HTML]{FFECF0}-} \\
\multicolumn{1}{c|}{\multirow{-4}{*}{TTA}} & \cellcolor[HTML]{C0C0C0}Ensemble & \cellcolor[HTML]{C0C0C0}\textbf{0.7741} & \cellcolor[HTML]{C0C0C0}0.7987 & \cellcolor[HTML]{C0C0C0}\textbf{0.7634} & \cellcolor[HTML]{C0C0C0}\textbf{0.9942} & \cellcolor[HTML]{C0C0C0}\textbf{0.6674} & \cellcolor[HTML]{C0C0C0}\textbf{0.7798} & \cellcolor[HTML]{C0C0C0}0.7734 & \cellcolor[HTML]{C0C0C0}\textbf{0.7989} & \cellcolor[HTML]{C0C0C0}\textbf{0.9945} & \cellcolor[HTML]{C0C0C0}\textbf{0.6772}
\end{tabular}%
}
\caption{Quantitative comparison of SwinV2-Tiny, DinoBloom-Small, ConvNeXt-Tiny, and their ensemble trained on two datasets: WBCBench Train + Eval, and the Expanded Dataset. Results are reported both with and without Test Time Augmentation (TTA). Metrics are computed from out-of-fold predictions to provide an unbiased evaluation; Test Macro F1 (highlighted in pink) serves as the held-out benchmark. Higher values indicate better performance; \textbf{Bold} indicates best performance and \underline{underline} indicates second-best. Our best model is the ensemble with TTA trained on the expanded dataset.}
\label{table:ablation}
\end{table*}
\section{Results}
We report macro F1-score, balanced accuracy, macro precision and macro specificity in Table \ref{table:ablation}. The metrics are computed based on out-of-fold predictions: for each fold $k \in \{1,2,3\}$, models are trained on $2$ folds and evaluated on the held-out fold $k$, such that each sample is evaluated exactly once by models not exposed to it during training, resulting in leakage-free logits for unbiased evaluation. We see that the ensemble of all three backbones trained on our expanded dataset performs the best across all metrics. We also show the out-of-fold confusion matrix for our ensemble model in Figure \ref{fig:confusion_matrices} right, which reveals an interesting biological insight: Myelocytes in the top-left of the matrix are commonly confused with each other, and the VLY and PLY classes are most commonly confused with lymphocytes. To emphasise this, we show the confusion matrix of  highly-confident misclassifications (i.e. the model gave the ground-truth class a very low probability) found through confident learning \cite{northcutt2021confidentlearning} in Figure \ref{fig:confusion_matrices} left. The dotted red box (top left) shows the granulopoiesis classes; excluding basophils and eosinophils, the solid red box shows that these classes are confused the most as biologically they represent the gradual maturation from a promyelocyte to a segmented neutrophil. The green box shows the classes in lymphopoiesis, which are also commonly confused with each other. Lastly, the confusion matrix reveals that cells labelled as `blast cells' are likely to be lymphoblasts and not myeloblasts, as they are confused along with LY and VLY. 

\section{Conclusion and Discussion}
We show that fine-tuning a pipeline of small pretrained classifiers achieves excellent performance on a highly imbalanced classification dataset of white blood cell images. Our approach ensembles $3$ distinct architectures --- self supervised DiNO, convolutional neural networks and hierarchical vision transformers --- in a stratified $3$-fold cross-validation framework. Our best score was $0.67726$ on the final competition test set. Despite this strong performance, we reason that errors were most prevalent around biologically similar classes along the myelocytes and lymphocytes and hypothesise that future work would benefit from designing specific experts for granulopoiesis, lymphopoiesis and monocytopoiesis to learn stronger inter-class representations and improve classification.

\section{Compliance with ethical standards}
\label{sec:ethics}
This is a numerical simulation study for which no ethical approval was required.

\section{Acknowledgments}
\label{sec:acknowledgments} 
We acknowledge \cite{wbcbench2026} for providing the WBCBench dataset. The authors acknowledge access to the Batch Compute System in the Department of Computer Science at the University of Warwick, and associated support services, in the completion of this work. T.B. was supported through EPSRC grant EP/V062522/1. J.B. is supported through EPSRC, project reference 2882348. T.B and A.S. are supported through EPSRC/NSF grant EP/X026663/1. S.B. is supported by Innovate UK through a Knowledge Transfer Partnership between the University of Warwick and Intelligent Imaging Innovations Ltd (Grant No. 10159795).

\bibliographystyle{IEEEbib}
\bibliography{refs}

\end{document}